\documentclass[letterpaper, 10 pt, conference]{ieeeconf}  

\IEEEoverridecommandlockouts                              

\usepackage{graphicx}
\usepackage{amsmath}
\usepackage{amssymb}
\usepackage{booktabs}

\title{\LARGE \bf
Confidence Optimization for Probabilistic Encoding
}

\author{Pengjiu Xia$^{1*}$, Yidian Huang$^{2}$, Wenchao Wei$^{1}$, Yuwen Tan$^{1}$    
\thanks{* Corresponding author. }
\thanks{
Pengjiu Xia$^{1}$, Wenchao Wei$^{1}$, and Yuwen Tan$^{1}$ are from the School of Computer Science and Technology, Beijing Institute of Technology, Beijing, 100081, China. 
Email: {\tt\small \{xiapengjiu, 1120210679, 3220241399\}@bit.edu.cn}.  
Yidian Huang$^{2}$  is from the School of Automation, Beijing Institute of Technology, Beijing, 100081, China. 
Email: {\tt\small hyd15213136303@gmail.com}.}
}

\begin{document}

\maketitle
\thispagestyle{empty}
\pagestyle{empty}

\begin{abstract}

Probabilistic encoding introduces Gaussian noise into neural networks, enabling a smooth transition from deterministic to uncertain states and enhancing generalization ability. However, the randomness of Gaussian noise distorts point-based distance measurements in classification tasks. To mitigate this issue, we propose a confidence optimization probabilistic encoding (CPE) method that improves distance reliability and enhances representation learning. Specifically, we refine probabilistic encoding with two key strategies: First, we introduce a confidence-aware mechanism to adjust distance calculations, ensuring consistency and reliability in probabilistic encoding classification tasks. Second, we replace the conventional KL divergence-based variance regularization, which relies on unreliable prior assumptions, with a simpler L2 regularization term to directly constrain variance. The method we proposed is model-agnostic, and extensive experiments on natural language classification tasks demonstrate that our method significantly improves performance and generalization on both the BERT and the RoBERTa model.

\end{abstract}

\section{INTRODUCTION}

\begin{figure}[b]
    \centering
    \includegraphics[width=0.88\linewidth]{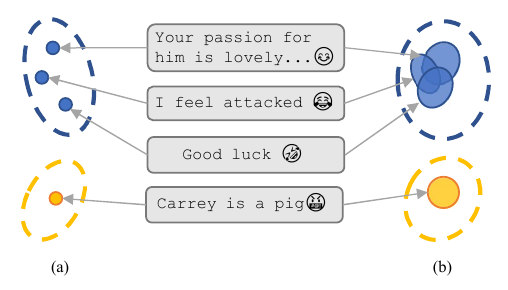}
    \caption{Taking the offense eval task as an example, deterministic encoding maps data to a single point like (a); probabilistic encoding maps data to a distribution. The same category will share the same feature space.}
    \label{fig:diffen}
\end{figure}

Probabilistic encoding, also known as uncertainty encoding, is a representation learning paradigm that maps deterministic feature vectors into probabilistic distributions\cite{bengio2013representation}, typically Gaussian, in a latent space. Unlike deterministic encoding, which represents data as fixed vectors, probabilistic encoding captures the inherent uncertainty in data, providing more prosperous and more flexible representations. By modeling features as distributions, it allows representations of the same category to form compact clusters while maintaining clear separability from other categories. The distribution of points in the feature space differs after deterministic encoding and probabilistic encoding, as illustrated in Fig.\ref{fig:diffen}. 

Probabilistic encoding has a long history in machine learning, with early applications in word vector representations\cite{vilnis2014word}. In computer vision, the Variational Autoencoder (VAE)\cite{kingma2013auto} introduces probabilistic encoding into generative models by mapping inputs into Gaussian distributions parameterized by mean and variance. This enables the generation of unseen samples. This work has a profound impact on
subsequent deep learning and generative model\cite{ho2020denoising}, inspiring subsequent advances in stochastic models. More recently, Hedged Instance Embeddings (HIB)\cite{oh2018modeling} extends probabilistic encoding to image retrieval and verification tasks. By encoding images as Gaussian distributions, HIB demonstrates improved robustness in tasks like handwritten digit recognition, especially on noisy or corrupted data. These results underscore the potential of probabilistic encoding in scenarios that require uncertainty estimation, motivating further exploration across various domains.

Probabilistic face embeddings (PFE)\cite{shi2019probabilistic} and Data Uncertainty Learning (DUL)\cite{chang2020data} are seminal works that demonstrate the power of probabilistic encoding in face recognition. PFE encodes each face as a latent Gaussian distribution, introducing the Mutual Likelihood Score (MLS) to measure the similarity between distributions. DUL further improves upon PFE by making both the mean and variance vectors learnable, achieving enhanced feature compactness and separability in the latent space. In addition to face recognition, probabilistic encoding has also been applied to multimodal retrieval, where data from different modalities are represented as distributions in a shared space\cite{chun2021probabilistic}, improving performance and interpretability. Recent advancements, such as Structured Probabilistic Coding (SPC)\cite{hu2024structured}, introduces structural regularization to learn more compact and enriched representations, showing promise in classification and regression tasks.

\begin{figure} [t]
    \centering
    \includegraphics[width=1 \linewidth]{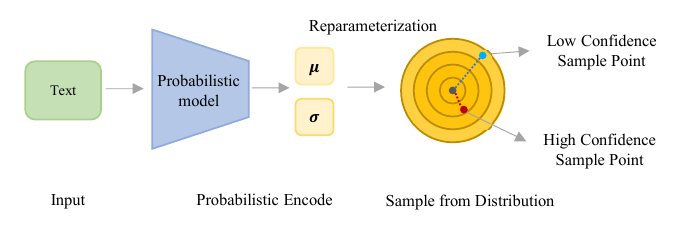}
    \caption{Encoding process of probabilistic models. Text input data is encoded as mean \( \mu_i \) and variance \( \sigma_i \) embeddings. The sampling process from the reparameterized distribution is stochastic, and the distance to the center point can be large or small. This distance reflects the confidence of the sampled points: points farther from the center have lower confidence, while those closer have higher confidence.}
    \label{fig:encode process}
\end{figure}

Despite these advancements, current probabilistic encoding methods face notable limitations, particularly in classification tasks where robustness and reliability are crucial. For instance, while probabilistic encoding effectively captures uncertainty by mapping features into latent distributions, distance measurement in classification still relies on individual sampled points as illustrated in Fig.\ref{fig:encode process}, which lacks robustness and necessary constraints. The reliance on a single sampled point overlooks the confidence associated with that point. When the point is far from the center of the distribution, the low confidence can lead to unreliable distance measures and poor classification performance. Conversely, points closer to the center, with higher confidence, lead to more stable and accurate results. This results in inaccurate feature distances in the latent space, increasing intra-class variance and reducing inter-class separability. Addressing these issues is crucial to fully harness the potential of probabilistic encoding in real-world applications.

Accordingly, we propose a confidence-based optimization method to enhance probabilistic encoding and further improve downstream performance. Our approach introduces additional constraints in the latent space, ensuring more consistent and reliable feature distances for downstream tasks. Specifically, we design a novel confidence-aware mechanism that mitigates the issue of overlapping feature distributions, improving both intra-class compactness and inter-class separability. Moreover, our method is model-agnostic and can be seamlessly integrated with various probabilistic encoding frameworks, making it broadly applicable across different architectures. The key contributions of our work are as follows:
\begin{itemize}
    \item \textbf{Novel confidence-aware mechanism}: We propose a normalized confidence metric to adjust distance calculations, ensuring consistency and reliability in probabilistic encoding-based classification tasks.
    \item \textbf{Improved regularization strategy}: We replace KL divergence with L2 regularization , avoiding using inaccurate prior assumptions and improving the learning process.
    \item \textbf{Enhanced downstream performance}: We conduct extensive experiments on benchmark datasets and demonstrate that our approach significantly improves classification accuracy and generalization, outperforming state-of-the-art probabilistic encoding methods.
\end{itemize}

\section{Preliminary}
\subsection{Problistic Encoding}
Representation learning, particularly through probabilistic encoding, has become a cornerstone of modern machine learning. Unlike deterministic approaches, probabilistic encoding models the latent space as a distribution rather than deterministic points, allowing better generalization and capturing data uncertainty. VAE model is a representative work of probabilistic encoding, which exemplifies this paradigm by learning the posterior distribution of latent variables given observed data.

VAE model consists of an encoder and a decoder. The encoder maps high-dimensional input data to the parameters of a low-dimensional latent distribution (e.g., mean and variance vectors), while the decoder reconstructs the data using samples from this latent distribution. In VAE model, the generated data is produced by a stochastic process. Let \( Z \) be the latent variable that is unobservable in the stochastic process, and \( X \) be the input data. The samples \( z_i \) and \( x_i \) are obtained from the random variables. The data generation process involves two steps: first, a latent variable \( z_i \) is sampled from the prior distribution \( P(Z) \) (typically \( \mathcal{N}(0, I) \)), and then a data point \( x_i \) is generated according to the conditional distribution \( P(X \mid Z) \). However, directly solving this process requires a large amount of sampling, which is complex and time-consuming, making it practically infeasible. Therefore, a posterior distribution \( P(Z \mid X) \) is introduced in the encoding part for joint modeling, representing the probability distribution of projecting the data into the latent space.

The posterior distribution \( P(Z \mid X) \) is sophisticated and intractable, but can be approximated using variational inference. Specifically, a simple, parameterized approximate distribution \( Q(Z \mid X) \) is introduced to approximate  \( P(Z\mid X) \). This approximation allows the VAE to bypass the direct computation of the intractable posterior, enabling efficient training through optimization of the Evidence Lower Bound (ELBO). The ELBO consists of two components: a reconstruction loss, typically measured by Mean Squared Error (MSE) or cross-entropy(CE), and a KL divergence regularization term. The KL divergence encourages the latent variable distribution to align with the predefined prior distribution \( \mathcal{N}(0, I) \) while preventing variance collapse \cite{asperti2019variational} during training, thereby enhancing the model's generalization ability.

\subsection{Downstream Tasks}
Our work focuses on natural language classification. Research on probabilistic encoding is abundant in the fields of computer vision and image processing. However, specific VAE methods are not commonly seen in the NLP domain\cite{FoK5nJNtXgRhtJa1}. Recently, SPC applied probabilistic encoding to NLP classification tasks, achieving better generalization. These findings suggest that probabilistic encoding has significant potential for text classification. 

Despite the success of the probabilistic encoding method in various tasks, current probabilistic encoding methods rely on fixed-point distance measurements for classification, leading to inaccurate distance calculation. Our CPE approach seeks to solve this problem by improving distance measurement reliability and enhancing performance in classification tasks.

\section{Methodology}
In this section, we provide a detailed introduction to the CPE method proposed in this paper, including the drawbacks of existing methods and our improvements. The reconstruction loss term, confidence loss term and regularization term, constitute our final optimization objective.
\subsection{Computation of Confidence}
Since deterministic modeling methods lack the ability to predict continuous functions and merely memorize the mapping relationships between discrete points, their predictive capability is limited. It is necessary to overcome the shortcomings of deterministic encoding models to enable them to predict continuous functions. This can be achieved using models such as VAE, which map discrete data points to a continuous latent feature space. Specifically, the representation \( z_i \) of each input \( x_i \) in the latent space follows a Gaussian distribution.
\begin{equation}
P(z_i \mid x_i) = \mathcal{N}(z_i \mid \mu_i, \sigma_i^2 I)
\label{eq:pzx}
\end{equation}
The mean \( \mu_i \) and variance \( \sigma_i \) of the Gaussian distribution are predicted by the mean and variance prediction network respectively. By introducing random noise, a reparameterization technique is used to sample from this distribution:

\begin{equation}
s_i = \mu_i + \epsilon \sigma_i; \quad \epsilon \sim \mathcal{N}(0, I)
\label{eq:si}
\end{equation}
This technique not only enables continuous predictions, but also ensures that gradients can flow through the sampling process, facilitating model optimization during training.
A key component of the optimization objective is the reconstruction loss, which is task-specific. For classification tasks, the commonly used cross-entropy loss is defined as:
\begin{equation}
L_{\text{CE}} = -\frac{1}{N} \sum_{i=1}^N \log \left( \frac{e^{w_{y_i}^T s_i}}{ \sum_{c} e^{w_c^T s_i}} \right)
\label{eq:LCE}
\end{equation}
Here, \(N\) is the number of all examples, \(w_c \) represents the weight vector for class \( c\), \( y_i\) is the true class label and \( s_i\) is the latent representation sampled. However, this approach has certain limitations. In classification tasks, the distances between points in the feature space are still computed deterministically. The uncertainty introduced by Gaussian noise may distort these distances, leading to unreliable results. For example, samples drawn near the center of the Gaussian distribution tend to have higher confidence, while those near the edges have lower confidence. These variations necessitate a method for assessing the reliability of probabilistic encoding. 

To solve this problem, we incorporate a confidence loss term into the optimization objective. This term evaluates and regularizes the confidence of the probabilistic encoding process, ensuring more reliable downstream task performance by mitigating the variability introduced during sampling.
For classification tasks, we incorporate the weight matrix \( W \in \mathbb{R}^{D \times C} \) from the classification head, where \( D\) and \(C\) denote the embedding dimension and the number of all classes, and \( w_c\) denotes the center of class \( c\). Inspired by the DUL, which introduced an uncertainty score \( \frac{(w_c - \mu)^2}{2 \sigma^2} \) to quantify uncertainty, we extend this concept to compute confidence levels. Unlike DUL, which neither includes uncertainty constraints in the optimization objective nor performs normalization, we redefine confidence using cumulative probability.

In our framework, discrete feature points are mapped to a Gaussian mixture distribution. The confidence of class \(c\) is intuitively represented by the cumulative probability of the Gaussian density, expressed as:

\begin{equation}
\text{density}_c = \frac{1}{\sqrt{2 \pi \sigma^2}} \exp\left(-\frac{(w_c - \mu)^2}{2 \sigma^2}\right)
\label{eq:density_c}
\end{equation}

\begin{align}
\text{confidence}_c &= \operatorname{Erf}\left(\frac{(w_c - \mu)^2}{2\sigma^2}\right) \notag\\
&= \frac{2}{\pi} \int_{0}^{\frac{(w_c - \mu)^2}{2\sigma^2}} e^{-t^2} \, dt
\label{eq:confidencec1}
\end{align}

Eq. (\ref{eq:density_c}) calculates the Gaussian probability density, while (\ref{eq:confidencec1}) integrates it to obtain the cumulative probability, representing the confidence level of class \(c\). Our objective includes maximizing confidence for relevant classes during training. However, directly applying (\ref{eq:confidencec1}) faces two challenges:
\begin{itemize}
    \item \textbf{Lack of normalization}: Without normalization across classes, multiple classes may simultaneously exhibit high confidence, reducing inter-class discriminability.
    \item \textbf{Deviation from prior assumptions}: The actual data distribution often deviates from the assumed standard normal distribution. 
\end{itemize}
For instance, features with large variance may result in near-uniform distributions, while small variance can collapse to deterministic encoding. This variability undermines the reliability of using the Gaussian Error Function(Erf). So we normalize confidence values across classes and feature dimensions as N-confidence. This ensures that the confidence is appropriately scaled, enhancing class separation and maintaining consistency with the actual feature distribution. The improved confidence formulation is expressed as:

\begin{equation}
\text{N-confidence}_c = \frac{e^{-\frac{(w_c - \mu)^2}{2\sigma^2}}}{\sum_{i} e^{-\frac{(w_i - \mu)^2}{2\sigma^2}}}
\label{eq:confidencec2}
\end{equation}

The numerator represents the Gaussian probability density of the current class, while the denominator represents the Gaussian probability densities of all classes or candidate classes. Since the numerator and denominator share common multiplicative factors, they have been omitted and simplified in (\ref{eq:confidencec2}). After obtaining the confidence level, we use maximum likelihood to represent the 
average confidence loss of all examples, as shown in (\ref{eq:Lconf}). Each sample \(x_i\) is mapped to a class \( i\) based on the real label.

\begin{align}
L_{\text{conf}} &= -\log\left(\frac{1}{N} \sum_{i=1}^{N} \text{N-confidence}_i\right)
\label{eq:Lconf}
\end{align}

When the distance \(|w_c - \mu|\) between the estimated class center and the mean is large, the variance representing uncertainty increases. Conversely, when the class center is closer to the mean, the model has higher certainty about these data, and the variance decreases. During confidence calculation, different feature dimensions impact classification outcomes differently, exhibiting positive correlation, negative correlation, or no correlation depending on the specific class. For correlated dimensions, maximizing confidence ensures that the class center approaches the distribution mean, effectively reducing uncertainty. However, overly high confidence can result in variance collapse, where the variance diminishes excessively. This collapse undermines the uncertainty of samples drawn from the Gaussian mixture, causing a degeneration into deterministic encoding and reducing the model's generalization capability.
\subsection{Solutions to Excessive Confidence}
In order to prevent variance collapse caused by excessively high confidence, we have made further optimizations based on the aforementioned confidence calculation, with improvements in two aspects.

On one hand, we optimize the classifier component by managing the impact of confidence on class center adjustments. High confidence can distort classification scoring, so we preemptively assess whether the confidence is excessively high. Specifically, we introduce two thresholds \( t_1 \) and \( t_2 \). The value of \( t_1 \)corresponds to a moderately confident score in the distribution, used to classify confidence values into high-confidence and low-confidence groups. The value of \( t_2 \) is set to 20\% of the aggregated confidence across all classes. We compare the lower bound \( \text{min}_v \) of the high-confidence group with the upper bound \( \text{max}_v \) of the low-confidence group. If the difference between \( \text{min}_v \) and \( \text{max}_v \) exceeds the threshold \( t_2 \), the confidence of the high-confidence group is considered excessively high. To mitigate this, we propose an Overly Mask, a binary mask matching the shape of the class centers \( W \), with all initial values set to 1. 
When the \(\text{N-confidence}_c\) of a dimension of the class center \( w_c \) exceeds a threshold \( t \), it is deemed excessively high, and the corresponding entry in the Overly Mask is set to 0. Before scoring with the classifier, the Overly Mask is applied to \( W \), masking dimensions with excessive confidence. Subsequently, classification loss is calculated, and adjustments are made to confidence values in these masked dimensions to prevent variance collapse.

On the other hand, VAE model employs the KL divergence as a regularization term to constrain the variance. The KL divergence regularization term measures the dissimilarity between two distributions. In VAE model, it is used to approximate the distribution of latent variables to the prior distribution, which is the standard normal distribution. This regularization serves as an information bottleneck, ensuring efficient latent encoding while maintaining generative flexibility. The specific calculation is shown in (\ref{eq:lkl}):
\begin{align}
L_{kl} &= \text{KL}\left[ \mathcal{N}(z_i; \mu_i, \sigma_i^2 I) \, \| \, \mathcal{N}(0, I) \right]  \notag\\
&= -\frac{1}{2} \left( 1 + \log \sigma_i^2 - \mu_i^2 - \sigma_i^2 \right)  
\label{eq:lkl}
\end{align}


However, different dimensions of features under the confidence constraint may not satisfy the prior assumption of a standard normal distribution, and using \( \mathcal{N}(0,I) \) as the prior for latent variables may not be accurate. Therefore, we use a simple and direct regularization approach: applying an L2 regularization term to the variance, as shown in (\ref{eq:ll2}). 

\begin{align}
    L_{l2} &= -\frac{1}{N} \sum_{i=1}^{N} \sigma_i^2
\label{eq:ll2}
\end{align}

Unlike the KL divergence regularization, L2 regularization directly optimizes the variance by encouraging its maximization, allowing for greater flexibility in feature representation. To validate the effectiveness of this approach, we conducted experiments comparing L2 variance regularization with KL divergence regularization. Results demonstrate that L2 regularization achieves comparable performance while better handling deviations from prior assumptions. The final optimization objective can be expressed as (\ref{eq:loss}). \( L_{norm} \) can be either \( L_{kl} \) or \( L_{l2} \). The parameters \( \lambda_1 \) and \( \lambda_2 \) represent the weights of the loss terms, determined using a hyperparameter search method for each task. In our experiments, we performed grid search tuning based on validation performance, selecting stable values that generalize well
across different datasets. The range of hyperparameter values is between 0.1 and 1. 

\begin{align}
    \text{\textit{Loss}} = L_{CE} + \lambda_1 \cdot L_{norm} + \lambda_2 \cdot L_{conf}
    \label{eq:loss}
\end{align}

\section{Experiments}
We evaluated our method on TweetEval Benchmark.

\subsection{TweetEval Benchmark}
In NLP system, the informal, fast-paced, and highly idiosyncratic nature of social media text presents unique challenges compared to more structured textual data. Sentiment analysis and other classification tasks are notably difficult due to the brevity of social media, the lack of sufficient contextual cues, and platform-specific constraints such as character limits. We refer to the work of Tweeteval Benchmark\cite{barbieri2020tweeteval} and SPC, selecting seven classification tasks in social media tweet analysis. These tasks are: \textbf{EmotionEval},\textbf{EmojiEval}, \textbf{IronyEval}, \textbf{HatEval}, \textbf{OffensEval}, \textbf{SentiEval}, and \textbf{StanceEval}.
The details of  Tweeteval Benchmark as Table \ref{tab:table}.

\begin{table}[t]
\centering
\caption{Number of instances in training, validation, and test sets}
\label{tab:table}
    \resizebox{0.48\textwidth}{!}{
    \large
    \begin{tabular}{*{10}{c}}
        \toprule
        \textbf{Task} & \textbf{Lab} & \textbf{Train} & \textbf{Val} & \textbf{Test} \\ 
        \midrule
        Emoji prediction & 20 & 45,000 & 5,000 & 50,000 \\ 
        Emotion recognition & 4 & 3,257 & 374 & 1,421 \\ 
        Hate speech detection & 2 & 9,000 & 1,000 & 2,970 \\ 
        Irony detection & 2 & 2,862 & 955 & 784 \\ 
        Offensive lg. id. & 2 & 11,916 & 1,324 & 860 \\ 
        Sentiment analysis & 3 & 45,389 & 2,000 & 11,906 \\ 
        Stance detection & 3 & 2,620 & 294 & 1,249 \\ 
        \bottomrule
        \bottomrule
        Stance / Abortion & 3 & 587 & 66 & 280 \\ 
        Stance / Atheism & 3 & 461 & 52 & 220 \\ 
        Stance / Climate & 3 & 355 & 40 & 169 \\ 
        Stance / Feminism & 3 & 597 & 67 & 285 \\ 
        Stance / H. Clinton & 3 & 620 & 69 & 295 \\ 
        \bottomrule
    \end{tabular}
    }
    \vspace{-1em}
\end{table}

\subsection{Implementation Details}
\newcommand{\meanpm}[2]{#1 \textstyle{\pm} \scriptstyle{#2}}
\begin{table*}[t]
    \centering
    \caption{Classifcation evaluation on 7 benchmark datasets.}
    \begin{tabular}{lcccccccc}
        \toprule
        \textbf{Methods}&\textbf{EmojiEval}&\textbf{EmotionEval}  &\textbf{HateEval}  &\textbf{IronyEval } &\textbf{OffensEval}  &\textbf{SentiEval } &\textbf{StanceEval}   &\textbf{Avg} \\
        \midrule
         SVM& 29.30 & 64.70 &36.70  &61.70  &52.30  &62.90  &67.30  &53.56 \\
         FastText&25.80  &65.20  &50.60  &63.10  &73.40  &62.90  &65.40  &58.06 \\
         BLSTM&24.70  &66.00  &52.60  &62.80  &71.70  &58.30  &59.40  &56.50 \\
        \midrule
            \multicolumn{2}{l}{\textbf{BERT Backbone}}\\
        \midrule
         CE&$\meanpm{22.30}{0.60}$ &$\meanpm{76.05}{1.41}$ &$\meanpm{44.67}{1.78}$ &$\meanpm{59.38}{3.01}$&$\meanpm{80.16}{1.26}$&$\meanpm{70.54}{0.44}$&$\meanpm{65.21}{0.71}$&59.76 \\
        MEIB&$\meanpm{21.87}{0.73}$&$\meanpm{76.70}{0.82}$&$\meanpm{48.27}{1.72}$&$\meanpm{65.87}{2.14}$&$\meanpm{80.49}{0.81}$&$\meanpm{70.55}{0.57}$ &$\meanpm{65.59}{1.58}$&61.33 \\
         SPC&$\meanpm{24.19}{1.55}$&$\meanpm{77.15}{0.73}$&$\meanpm{57.48}{2.99}$&$\meanpm{65.85}{1.07}$&$\meanpm{\textbf{80.65}}{0.78}$&$\meanpm{\textbf{70.74}}{0.12}$ &$\meanpm{\textbf{67.17}}{1.08}$&63.32 \\
         CPE(ours)&$\meanpm{\textbf{26.00}}{0.18}$ &$\meanpm{\textbf{78.13}}{0.27}$ &$\meanpm{\textbf{62.04}}{1.90}$&$\meanpm{\textbf{68.53}}{1.07}$&$\meanpm{80.61}{0.17}$&$\meanpm{70.45}{0.12}$ & $\meanpm{67.02}{0.33}$ &\textbf{64.68} \\
        \midrule
            \multicolumn{2}{l}{\textbf{RoBERTa Backbone}}\\
        \midrule
         CE&$\meanpm{30.25}{1.32}$&$\meanpm{77.41}{1.33}$ &$\meanpm{45.49}{4.70}$ &$\meanpm{57.99}{4.96}$ &$\meanpm{78.74}{2.20}$ &$\meanpm{71.80}{0.93}$  &$\meanpm{66.78}{1.34}$ &61.21 \\
        MEIB&$\meanpm{29.94}{1.30}$&$\meanpm{78.73}{0.90}$&$\meanpm{49.34}{2.42}$&$\meanpm{60.54}{2.70}$&$\meanpm{79.68}{0.98}$&$\meanpm{72.78}{0.29}$ &$\meanpm{67.89}{1.70}$&62.70 \\
         SPC&$\meanpm{32.54}{0.48}$ &$\meanpm{79.01}{0.61}$ &$\meanpm{\textbf{59.80}}{1.32}$ &$\meanpm{65.31}{1.91}$ &$\meanpm{\textbf{80.98}}{1.36}$ &$\meanpm{\textbf{72.96}}{0.22}$ &$\meanpm{69.02}{0.63}$ &65.65 \\
         CPE(ours)& $\meanpm{\textbf{33.71}}{0.30}$& $\meanpm{\textbf{79.47}}{0.23}$ &$\meanpm{58.26}{1.10}$&$\meanpm{\textbf{67.50}}{0.80}$  &$\meanpm{80.63}{0.55}$&$\meanpm{72.93}{0.12}$ & $\meanpm{\textbf{69.10}}{0.41}$ &\textbf{65.94} \\

        \bottomrule
    \end{tabular}

    \vspace{0.4em}
    \small  
    \parbox{\dimexpr\linewidth-8\tabcolsep}{ 
        \raggedright 
        \textbf{Note:} 
        The results of the SVM, FastText, and BLSTM methods come fromBarbieri, while the results for CE, MElB, and SPC come from SPC. The best results for each benchmark are highlighted inbold. Following previous experiments, our experiments also used 5 random seeds, and the results are averaged for comparison.
    }

    \label{tab:nlp result}
\end{table*}

\begin{table*}[t]
    \centering
    \caption{Results of ablation experiment, RoBERTa as the backbone.}
    \begin{tabular}{lcccccccc}
        \toprule
        \textbf{Methods}&\textbf{EmojiEval}&\textbf{EmotionEval}  &\textbf{HateEval}  &\textbf{IronyEval } &\textbf{OffensEval}  &\textbf{SentiEval } &\textbf{StanceEval}   &\textbf{Avg} \\
        \midrule
         CPE(ours)& $\meanpm{\textbf{33.71}}{0.30}$& $\meanpm{79.47}{0.23}$ &$\meanpm{\textbf{58.26}}{1.10}$&$\meanpm{67.50}{0.80}$  &$\meanpm{80.63}{0.55}$&$\meanpm{\textbf{72.93}}{0.12}$ & $\meanpm{\textbf{69.10}}{0.41}$ &\textbf{65.94} \\
        CPE$_{KL}$&$\meanpm{33.21}{0.21}$&$\meanpm{\textbf{79.56}}{0.21}$&$\meanpm{55.72}{1.42}$&$\meanpm{\textbf{67.64}}{1.00}$&$\meanpm{80.54}{0.19}$&$\meanpm{72.91}{0.17}$ &$\meanpm{69.08}{0.46}$&65.52 \\
        - w/o Norm&$\meanpm{33.17}{0.36}$&$\meanpm{79.43}{0.12}$&$\meanpm{55.29}{2.19}$&$\meanpm{66.87}{0.73}$&$\meanpm{\textbf{80.86}}{0.12}$&$\meanpm{72.88}{0.18}$ &$\meanpm{69.10}{0.45}$&65.37 \\
        - w/o Conf &$\meanpm{33.08}{0.37}$ &$\meanpm{79.32}{0.28}$ &$\meanpm{55.91}{1.81}$&$\meanpm{67.18}{1.22}$&$\meanpm{80.81}{0.25}$&$\meanpm{72.92}{0.23}$ & $\meanpm{69.18}{0.59}$ &65.48 \\
         CE&$\meanpm{30.25}{1.32}$&$\meanpm{77.41}{1.33}$ &$\meanpm{45.49}{4.70}$ &$\meanpm{57.99}{4.96}$ &$\meanpm{78.74}{2.20}$ &$\meanpm{71.80}{0.93}$  &$\meanpm{66.78}{1.34}$ &61.21 \\
         \midrule
         SPC&$\meanpm{32.54}{0.48}$ &$\meanpm{79.01}{0.61}$ &$\meanpm{\textbf{59.80}}{1.32}$ &$\meanpm{65.31}{1.91}$ &$\meanpm{\textbf{80.98}}{1.36}$ &$\meanpm{72.96}{0.22}$ &$\meanpm{69.02}{0.63}$ &65.65 \\
         - w Conf &$\meanpm{\textbf{33.49}}{0.11}$ &$\meanpm{\textbf{79.26}}{0.35}$ &$\meanpm{58.47}{2.24}$&$\meanpm{\textbf{66.77}}{1.61}$&$\meanpm{80.07}{0.75}$&$\meanpm{\textbf{73.08}}{0.07}$ & $\meanpm{\textbf{69.82}}{0.79}$ &\textbf{65.85} \\
        \bottomrule
    \end{tabular}
    \vspace{0.4em}
    
    \small  
    \parbox{\dimexpr\linewidth-8\tabcolsep}{ 
        \textbf{Note:} 
        CPE$_{KL}$ indicates replacing the regularization term with KL divergence. w/o Norm means removing the regularization term. w/o Confi means removing the confidence loss term. w Confi means adding the confidence loss term.
    }
    \label{tab:ablation result}
    
\end{table*}
Experiments were conducted on NVIDIA RTX 2080Ti GPUs. The batch size used in our experiments is 128. We employed the AdamX optimizer with a learning rate of 5e-5. The model was trained for a total of 20 epochs to ensure sufficient convergence. Additionally, we set the maximum sequence length of the tokenizer to 128 to balance computational efficiency and performance. These hyperparameter choices were made based on empirical evaluations and best practices to achieve optimal training stability and generalization.

In this study, we focus on the optimization role of confidence in probabilistic encoding rather than the encoder architecture. Our approach is model-agnostic, meaning it does not rely on specific architectural designs or modifications to the encoder and can be seamlessly integrated into various probabilistic encoding frameworks without being restricted to a particular model structure. For natural language classification, we use the BERT\cite{DBLP:journals/corr/abs-1810-04805} and RoBERTa\cite{liu2019roberta} base size models to encode text into 768-dimensional feature vectors. Both tasks use a Softmax classifier with weight W.

\subsection{Evaluation Metrics}
In Tweet Eval tasks, we follow the evaluation metrics from previous works and adopt macro-F1 as the primary evaluation metric and apply task-specific metrics where appropriate. Specifically, for StanceEval, we average the F1 scores of the against and favor categories; for IronyEval, we use the F1 score for the ironic category; and for the Sentiment task, recall is used as the metric. We compared the test results of the proposed confidence optimization method CPE with SVM\cite{cortes1995support}, BiLSTM\cite{hochreiter1997long}, FastText method\cite{joulin2016bag} specialized for NLP tasks, the probability encoding-based MEIB\cite{an2023maximum}, the SPC and the CE method which uses only cross-entropy.

\section{Overrall Results}

\subsection{Performance on Tweet Eval Benchmark}
Table.\ref{tab:nlp result} summarizes the performance of our method across 7 NLP classification benchmarks. Using both BERT and RoBERTa backbones, our CPE method achieves average performance improvements of \textbf{4.92\%} and \textbf{4.73\%} compared to the CE method respectively. Overall, our method delivers the best classification performance on most tasks. Notably, the BERT backbone achieves better results for HateEval and IronyEval, while the RoBERTa backbone demonstrates superior overall average performance across the benchmarks, highlighting its robustness for diverse TweetEval tasks. Whether using the probability-based encoding BERT model or the RoBERTa model, our approach consistently improves performance, demonstrating that our method is model-agnostic.

In addition, we evaluated the impact of our proposed optimization method on computational efficiency. We measured the training time before and after applying our optimization on tweet stance eval task. The overall training time overhead increased by only  \textbf{0.9\%} compared to the pre-optimization stag. This minor computation cost increase comes with centain benefits in terms of performance improvements.

\subsection{Ablation Experiment}
\addtolength{\textheight}{-0.5cm}   
We conduct several ablation experiments on NLP classification tasks, including removing the confidence loss term (w/o Confidence), comparing the performance of different regularization terms by replacing the L2 regularization term with KL divergence regularization, and the results of experiments with the regularization term removed (w/o Norm). Additionally, we test the effect of adding the confidence loss term on SPC, with the overall results shown in Table.\ref{tab:ablation result}

After replacing the L2 regularization term with the KL divergence term, the average metrics decrease, and performance on certain benchmarks is lower than with L2 regularization. This indicates that L2 regularization is more effective in this scenario, whereas KL divergence may introduce unreliable prior assumption. When the regularization term is removed, the average metrics also decrease, highlighting the crucial role of regularization in the optimization process. Similarly, removing the confidence loss term and retaining only the cross-entropy and KL divergence terms leads to performance degradation, with results lower than both our CPE method and the original SPC method. Adding the confidence loss term to the SPC method also leads to performance improvements, further validating the effectiveness of our confidence optimization method.


\section{Conclusion}
In this paper, we propose a Confidence Optimization for Probabilistic Encoding (CPE), a model-agnostic approach, to improve inaccurate distance measurements in downstream tasks with probabilistic encoding. Our approach improves the reliability of distance calculations in classification tasks, thus improving representation learning and classification performance. To mitigate variance collapse caused by excessively high confidence, we take the following solutions: Recognizing that the actual data distribution may deviate from the assumed standard normal prior, we replace the KL divergence regularization term with an L2 regularization term on the variance to ensure stability and prevent collapse. Our method achieves performance comparable to or better than previous SOTA models in multiple tasks. Extensive experiments on TweetEval benchmarks demonstrate that our method effectively learns uncertainty representations, improves classification accuracy, and exhibits strong generalization capabilities without compromising computational efficiency.




\section*{ACKNOWLEDGMENT}
We are grateful to our team members and advisor for their insightful discussions on technical details, which have significantly improved the quality of this research.


\bibliographystyle{IEEEbib}
\bibliography{icme2025references}
\end{document}